\documentclass[conference]{IEEEtran}
\usepackage{spconf,amsmath,graphicx}
\usepackage{booktabs}
\usepackage{graphicx}
\usepackage{multirow}
\usepackage{amsmath}
\usepackage{amssymb}
\usepackage{xcolor}
\usepackage{threeparttable}

\title{Proactive Adversarial Defense: Harnessing Prompt Tuning in Vision-Language Models to Detect Unseen Backdoored Images}
\name{Kyle Stein$^1$, Andrew A. Mahyari$^{1,2}$, Guillermo Francia, III$^3$, Eman El-Sheikh$^3$}
\address{$^1$ Department of Intelligent Systems and Robotics, University of West Florida, Pensacola, FL, USA\\
$^2$ Florida Institute For Human and Machine Cognition (IHMC), Pensacola, FL, USA \\
$^3$ Center for Cybersecurity, University of West Florida, Pensacola, FL, USA}

\begin{document}
\maketitle

\begin{abstract}
Backdoor attacks pose a critical threat by embedding hidden triggers into inputs, causing models to misclassify them into target labels. While extensive research has focused on mitigating these attacks in object recognition models through weight fine-tuning, much less attention has been given to detecting backdoored samples directly. Given the vast datasets used in training, manual inspection for backdoor triggers is impractical, and even state-of-the-art defense mechanisms fail to fully neutralize their impact. To address this gap, we introduce a groundbreaking method to detect unseen backdoored images during both training and inference. Leveraging the transformative success of prompt tuning in Vision Language Models (VLMs), our approach trains learnable text prompts to differentiate clean images from those with hidden backdoor triggers. Experiments demonstrate the exceptional efficacy of this method, achieving an impressive average accuracy of 86\%  across two renowned datasets for detecting unseen backdoor triggers, establishing a new standard in backdoor defense.
\end{abstract}


%
\IEEEpeerreviewmaketitle

\section{Introduction}

\IEEEPARstart{D}{eep} neural networks (DNNs) have revolutionized fields ranging from object classification \cite{li2022mvitv2} and face recognition \cite{boutros2022elasticface} to reinforcement learning \cite{mahyari2021policy} and natural language processing \cite{devlin2018bert}, setting new benchmarks in performance and innovation. However, this remarkable success has made them prime targets for sophisticated adversarial manipulations. Among the most insidious threats are backdoor attacks, which stealthily embed hidden patterns—known as triggers—into models, causing them to misclassify inputs into an adversary's chosen target label. These backdoors can be implanted through malicious techniques like data poisoning \cite{liu2020reflection} or neuron hijacking \cite{liu2018trojaning}, posing an immediate and formidable challenge. In response, the research community has developed numerous defense and detection strategies \cite{song2020universal, singla2022minimal, shi2024black, yuan2023activation, wei2024shared}. Early approaches focused on purifying compromised models using methods such as fine-tuning \cite{chen2021refit, weber2023rab} or distillation \cite{li2021neural}. More recently, cutting-edge techniques have attempted to neutralize adversarial triggers by leveraging limited training or in-distribution samples \cite{shi2024black, chai2022one, lee2018simple}. Another method adopts an input-level perspective by scaling pixel intensities of an image and checking consistency in the model’s predictions \cite{guo2023scale}. Furthermore, researchers in \cite{mumcu2024fast} utilize a Vision Transformer (ViT) to classify previously seen adversarial attack patterns targeting traffic sign recognition systems in autonomous vehicles. These advances mark significant strides in safeguarding DNNs, but the persistence and evolution of adversarial threats demand continued innovation to stay ahead in this escalating arms race.


\begin{figure*}[t!]
\centering
\includegraphics[scale=0.32]{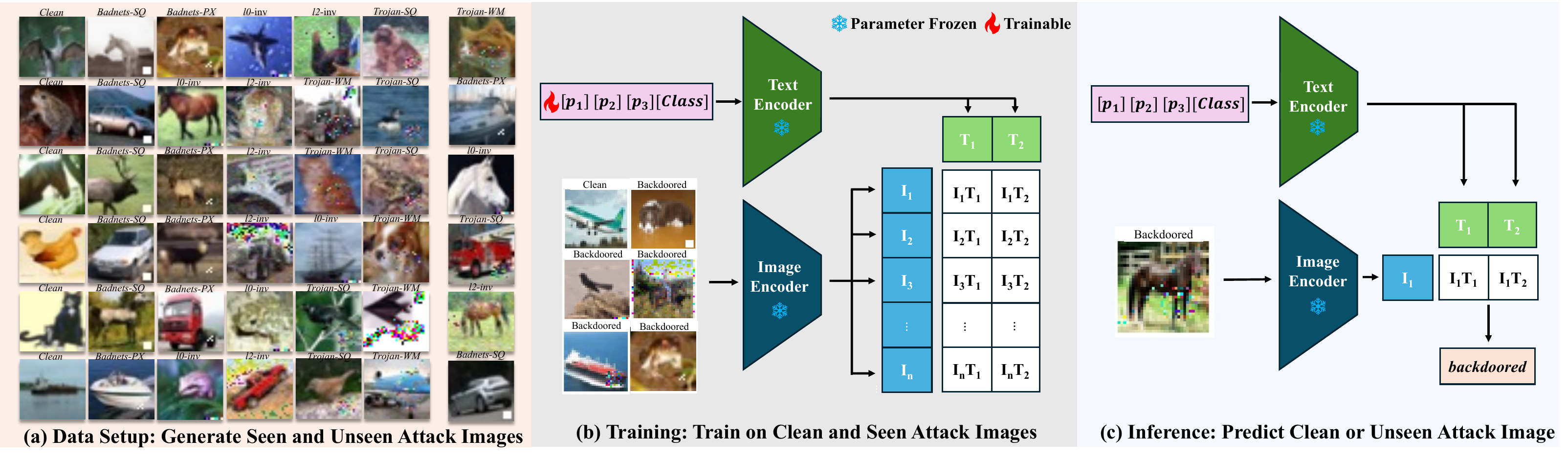}
\caption{The overall architecture of the proposed method on classifying unseen backdoor attack images. 
}
\label{fig:architecture}
\end{figure*}

\noindent \textbf{Contribution:} Despite advancements in adversarial defense and training algorithms, achieving 100\% protection against adversarial attacks remains elusive. These traditional methods are reactive, focusing on cleansing already-compromised models of embedded backdoor triggers. 
\textbf{Unlike the state-of-the-art methods, such as BDetCLIP \cite{niu2024bdetclip}, which utilize a poisoned model to uncover adversarial samples aimed at compromising the target model, our approach uses a clean and not poisoned CLIP model to detect unseen, open-world adversarial samples without having any prior knowledge about the attacks. This step is critical as millions of training samples are collected from the internet to train GenAI with little assurance that these samples are free from adversarial attacks.} This paper introduces a revolutionary and complementary strategy: a proactive algorithm designed to detect adversarial images before they wreak havoc. Our approach serves two critical purposes: \textit{i) Pre-Training Defense:} Before training begins, the algorithm meticulously scans the dataset to identify and eliminate adversarial (backdoored) images that could poison object recognition models. This ensures the integrity and purity of the training data, safeguarding the foundation of model learning. \textit{2) Inference-Time Shielding:} During inference, the algorithm acts as a vigilant gatekeeper, inspecting incoming images to block adversarial content from reaching the object recognition system. This prevents adversarial images from manipulating the model to misclassify inputs into the adversary's target class. By proactively identifying and neutralizing adversarial threats, this approach works in tandem with traditional defense algorithms, creating a robust and comprehensive safeguard against adversarial attacks. It represents a significant leap forward in securing object recognition systems.

\section{Proposed Method}
\label{sec:proposed-method}


\subsection{Preliminaries and Insignts}
\label{sec:preliminary}


Let $\mathcal{D}=\{ (x_i, y_i) \}_{i=1}^N$ represent a benign training set with $N$ images, where $x_i \in \mathcal{X}$ is the $i$-th image and $y_i \in \mathcal{Y} = \{1, \ldots,K\}$ is its corresponding label, with $K$ being the number of classes. 
If the goal of adversaries is to poison object recognition systems, they generate a poisoned dataset $\mathcal{D}_p$ to train the attacked model using either a standard loss function or an adversary-specified one. Specifically, $\mathcal{D}_p$ is composed of a modified subset of $\mathcal{D}$, denoted as $\mathcal{D}_m$, and a benign subset $\mathcal{D}_b$, such that $\mathcal{D}_p = \mathcal{D}_m \cup \mathcal{D}_b$, where $\mathcal{D}_b \subset \mathcal{D}$. The modified subset $\mathcal{D}_m$ is defined as $\mathcal{D}_m = \{ (x', y') \ | \ x' = G_X(x), y' = G_Y(y), (x,y) \in \mathcal{D}_s \}$, where $\alpha = \frac{|\mathcal{D}_s|}{|\mathcal{D}|}$ is the poisoning rate, $\mathcal{D}_s$ is the subset of $\mathcal{D}$ selected to be modified, and $G_X, G_Y$ are the adversary-specified generators for poisoned images and labels, respectively. For example, in BadNets~\cite{gu2019badnets}, $G_X(x) = (1-\beta) \odot x + \beta \odot t$, where {$\beta \in \{0,1\}^{C \times W \times H}$}, where $C, W$ and $H$ are dimensions of input,  $t$ is the trigger pattern, and $\odot$ denotes the element-wise product. 
If the goal of advesaries is to attack object recognition systems during inference, given an 'unseen' image $\hat{x}$ with ground-truth label $\hat{y}$, the model will predict its poisoned version $G_X(\hat{x})$ as $G_Y(\hat{y})$. \textbf{It is important to note that the goal of this paper is not to train any model or defend against attacked models, but to detect backdoored images $G_X(x)$ and $G_X(\hat{x})$.}

\noindent \textbf{VLM.} To detect unseen backdoored images, we harness the extraordinary capabilities of VLMs, which excel at transforming input images into lower-dimensional, highly informative feature spaces. Models like Contrastive Language-Image Pre-training (CLIP) \cite{radford2021learning}, trained on an immense dataset of image-text pairs (\textit{i.e. 400 million samples}), are uniquely equipped to extract rich and versatile features from input images, regardless of their intended application. Their joint training with text data makes VLMs particularly well-suited for advanced techniques like prompt tuning. Prompt tuning has emerged as a groundbreaking method in leveraging VLMs for diverse applications. In this process, a learnable soft prompt is passed through the language encoder of the VLM while keeping the weights of both the image and language encoders frozen. This innovative approach tailors VLMs—originally trained in an unsupervised manner on broad image-text data—to excel in specific downstream tasks. VLMs have a word embedding module $E(\cdot)$ that converts the symbolic words to word embeddings, a text encoder $f_t(\cdot)$ that maps the word embeddings to the joint space, and an image encoder $f_I(\cdot)$ that maps input images to the joint space.

\begin{table*}[t!]
\centering
\caption{Experimental Results of Unseen Attack Classification (Accuracy).}
\label{tab:main_results}
\scriptsize
\begin{tabular}{|l|l|l|l|l|l|l|l|l|}
\hline
\textbf{Dataset} & \textbf{Method} & \textbf{Trojan-WM} & \textbf{Trojan-SQ} & \textbf{$l_2$-inv} & \textbf{$l_0$-inv} & \textbf{Badnets-SQ} & \textbf{Badnets-PX} & \textbf{Average} \\ \hline
\multirow{3}{*}{CIFAR-10} 
& Simple-CNN \cite{eykholt2018robust} & 64.86 $\pm$ 7.14 & 75.10 $\pm$ 11.98 & 51.56 $\pm$ 0.41 & 49.93 $\pm$ 0.50 & 49.94 $\pm$ 0.42 & 50.10 $\pm$ 0.15 & 56.92 $\pm$ 3.43 \\ 
& Deep-CNN \cite{stallkamp2012man} & 76.37 $\pm$ 6.20 & 58.69 $\pm$ 6.72 & 55.77 $\pm$ 5.77 & 50.18 $\pm$ 0.34 & 50.14 $\pm$ 0.40 & 50.04 $\pm$ 0.06  &  56.87 $\pm$ 3.25\\ 
& ResNet-18 \cite{he2016deep} & 84.52 $\pm$ 5.22 & 75.40 $\pm$ 7.22 & 53.42 $\pm$ 1.28 &  51.17 $\pm$ 1.38 & 50.00 $\pm$ 0.00 & 54.29 $\pm$ 5.73  & 61.47 $\pm$ 3.47\\ 
& Proposed Method & \textbf{96.10 $\pm$ 0.38} & \textbf{96.44 $\pm$ 0.45} & \textbf{93.85 $\pm$ 0.69} & \textbf{96.45 $\pm$ 0.44} & \textbf{75.45 $\pm$ 2.03}  & \textbf{58.89 $\pm$ 1.31} & \textbf{86.20 $\pm$ 0.88}\\ \hline
\multirow{3}{*}{GTSRB} 
& Simple-CNN \cite{eykholt2018robust} & 82.86 $\pm$ 12.15 & 60.80 $\pm$ 10.35 & 53.58 $\pm$ 0.14 & 50.84 $\pm$ 0.30 & 50.00  $\pm$ 0.00 & 50.03 $\pm$ 0.03 & 58.02 $\pm$ 3.83\\ 
& Deep-CNN \cite{stallkamp2012man} & 83.93 $\pm$ 14.91 & 61.61 $\pm$ 13.95 & 53.17 $\pm$ 0.20 & 51.45 $\pm$ 0.47 & 50.01 $\pm$ 0.20 & 50.04 $\pm$ 0.10 & 58.36 $\pm$ 4.97\\ 
& ResNet-18 \cite{he2016deep} & 74.40 $\pm$ 5.93 & 87.52 $\pm$ 6.81 & 52.75 $\pm$ 1.39 & 69.29 $\pm$ 14.85 & 50.01 $\pm$ 0.01 & 50.02 $\pm$ 0.05  &  64.00 $\pm$ 4.84\\ 
& Proposed Method & \textbf{94.89 $\pm$ 0.78} & \textbf{95.72 $\pm$ 0.69} & \textbf{94.41 $\pm$ 0.65} & \textbf{86.99 $\pm$ 0.62} & \textbf{85.03 $\pm$ 0.62} & \textbf{60.42 $\pm$ 1.19} & \textbf{86.24 $\pm$ 0.76}\\ \hline
\multicolumn{9}{l}{\scriptsize \textbf{Note:} The values represent mean $\pm$ standard deviation over three random seeds. Bold indicates the best accuracy results for the unseen attack.} \\
\end{tabular}
\end{table*}

\subsection{Backdoor Image Detection Framework}

The framework of our proposed method for detecting unseen backdoored images is illustrated in Figure~\ref{fig:architecture}, showcasing a novel and powerful approach to tackling the complex challenge of detecting unseen backdoored attacks images. We leveraged prompt tuning of VLMs to transcend their general-purpose design to become specialized tools for detecting and classifying backdoored images with unparalleled precision and adaptability \cite{shen2021much, zhou2022learning, zhou2022conditional}. Figure~\ref{fig:architecture}(a) highlights the diverse permutations of backdoor attack images generated across the CIFAR-10 dataset, emphasizing the robustness of our method in handling a wide range of malicious image manipulations. Unlike traditional defenses focused on protecting against backdoored models, our goal is to directly detect backdoored images, combining training and test images into a unified dataset for this purpose.

In the training phase, depicted in Figure~\ref{fig:architecture}(b), we leverage the state-of-the-art architecture of CLIP, utilizing both a text encoder and an image encoder to project prompts and input images into a shared, semantically rich embedding space. Learnable soft prompts $[p_1, p_2, p_3]$ are prepended to the word embedding of the ``class'' label, which is either clean for normal images or backdoored for malicious ones. To speed up the convergence time, we initialize $[p_1, p_2, p_3]$ with the word embeddings of ``a photo of'', where \( p_i \in \mathbb{R}^d \) and $d$ is the dimension of the output of CLIP's word embedding $E(\cdot)$, and is equal to 512. This sequence, combining the learnable soft prompts and the word embedding of the ``class,'' is passed through the text encoder of CLIP and normalized, producing embedding vectors $T_1=f_t([p_1, p_2, p_3, E(\text{`clean'})])/\|f_t([p_1, p_2, p_3, E(\text{`clean'})])\|$ for clean images and $T_2=\frac{f_t([p_1, p_2, p_3, E(\text{`backdoored'})])}{\|f_t([p_1, p_2, p_3, E(\text{`backdoored'})])\|}$ for backdoored images. These embeddings capture the contextual nuances of the respective image classes.

Simultaneously, the image encoder processes all clean and backdoored images to generate high-dimensional embeddings $(I_1, I_2, \dots, I_n)$, where $I_i=\frac{f_I(x_i)}{\|f_I(x_i)\|}$. To optimize the model, similarity scores $(I_j \times T_1, I_j \times T_2)$ are computed between the $j$th image embedding and text embeddings $T_1$ and $T_2$, and a cross-entropy loss function is employed to fine-tune the system. This comprehensive framework not only ensures precise detection of seen backdoored images but also paves the way for identifying unseen backdoor attacks with exceptional accuracy and adaptability.


During inference, as shown in Fig.~1(c), the learned prefix embeddings $[p_1, p_2, p_3]$ are appended to the word embeddings of "clean" and "backdoored" and passed through the text encoder to generate the frozen text embeddings: $T_1 = f_t([p_1, p_2, p_3, E(\text{`clean'})])$, and $T_2 = f_t([p_1, p_2, p_3, E(\text{`backdoored'})])$. It is important to highlight that, although this process resembles the training phase, the prefix embeddings $p_1$, $p_2$, and $p_3$ are frozen during inference and remain unaltered. Meanwhile, the image encoder processes the input image $x_j$ to compute its corresponding embedding $I_j$. Finally, the similarity scores between $I_j$ and $T_1$, as well as $I_j$ and $T_2$, are calculated and compared: $\text{Similarity}(I_j, T_1), \quad \text{Similarity}(I_j, T_2)$. These similarity scores determine whether $x_j$ is classified as clean or backdoored. This architecture provides robust detection of unseen backdoored images by aligning embeddings of clean and adversarial images with their corresponding text embeddings in a shared multimodal space.

\noindent \textbf{Training.} During training, the model optimizes the alignment between the fixed visual embeddings, $I_j$, and the learnable text embeddings, $T_k, k \in \{1,2\}$ , to enable recognition of adversarial images. For each image $x_j$, the similarity scores for each class \( c \in \{\text{``clean''}, \text{``backdoored''}\} \) are calculated using the scaled dot product \( s_{j,k} = \alpha \times (I_j \cdot T_k) \), where \( \alpha \) is a scaling factor that amplifies the logits, and \( (\cdot) \) denotes the dot product operation. The scaling factor ensures that the logits are in a range suitable for the cross-entropy loss function.



The similarity scores $s_{j,k}$ are passed to the cross-entropy loss function, which encourages the model to assign higher similarity scores to the correct class. The cross-entropy loss function is defined as \( L = -\frac{1}{N} \sum_{j=1}^{N} \log \left( \frac{\exp(s_{j,k})}{\exp(s_{j,1})+\exp(s_{j,2})} \right) \), where \( N \) is the batch size, \( k \) is the true class label for the \( j \)-th sample, and \( s_{j,k} \) represents similarity scores for all classes \( k \) for sample \( j \). The learnable prefix embeddings in the text encoder are optimized using the Adam optimizer \cite{kingma2014adam}.

\noindent \textbf{Inference.} During inference, for each input image $x_j$, we compute the similarity scores $s_{j,k}$ for each class $k$. However, instead of computing class probabilities using the softmax function, we select the class with the highest similarity score, resulting in the predicted class label \( \hat{k}\) given by \( \hat{k} = \arg\max_{k} s_{j,k} \). Notably, the model is tested on unseen attack images, leveraging the information learned from seen attacks during training to generalize to novel unseen images.

\section{Experiments}
\label{sec:experiment}



\noindent \textbf{Attack Models.} We have selected six renowned backdoor attacks to evaluate our proposed architecture: Badnets Square (Badnets-SQ) \cite{gu2019badnets}, Badnets Pixels (Badnets-PX) \cite{gu2019badnets}, Trojan Square (Trojan-SQ) \cite{liu2018trojaning}, Trojan Watermark (Trojan-WM) \cite{liu2018trojaning}, $l_2$-inv \cite{li2020invisible}, and $l_0$-inv \cite{li2020invisible}. These attacks cover a wide range of backdoor conditions, including universality, label specificity, and variations in backdoor shape, size, and location.  




\noindent \textbf{Datasets.} We conduct experiments using two datasets: CIFAR-10 \cite{krizhevsky2009learning} and GTSRB \cite{stallkamp2012man}. CIFAR-10 includes 50,000 training images and 10,000 test images across 10 classes. GTSRB consists of 39,209 training images and 12,630 test images of traffic signals, spanning 43 classes.

\noindent \textbf{Experiment Settings.} It is important to note that unlike adversarial attacks and defense literature that work on the model, we are working on images solely to detect backdoored images. Therefore, we do not train any model (e.g. ResNet-18) for our evaluation. We use CLIP's ViT-B/32, the smallest architecture in the CLIP family, chosen for its efficiency in balancing performance and computational demands. We leverage the predefined training and testing splits from the previously mentioned datasets. Clean images are taken directly from the train and test split without modification, while backdoored images are generated by applying the predefined attack types to all clean images. To evaluate the performance of our proposed method in detecting \textbf{unseen attacks}, only five of six attack types are selected for training. To maintain balance, we randomly select an equal number of images from each attack type to match the total number of clean images. At inference, the model is tested on all clean test images and their backdoored version by the \textbf{unseen attack}, which is the attack type excluded during training. Furthermore, we set the scaling factor \( \alpha \) = 100 and the Adam optimizer's learning rate for the learnable prefix embeddings to $10^{-5}$. Training is conducted over 10 epochs with a batch size of 128.

\begin{table}[t!]
\centering
\scriptsize
\begin{threeparttable}
\caption{Cross-Generalization Results (Accuracy).}
\label{tab:cross-gen}
\begin{tabular}{|l|c|c|}
\hline
\textbf{Unseen Attack} & \textbf{CIFAR-10 $\rightarrow$ GTSRB} & \textbf{GTSRB $\rightarrow$ CIFAR-10} \\ \hline
Trojan-WM       &   76.54 $\pm$ 0.97                    &   80.24 $\pm$ 0.83       \\ 
Trojan-SQ       &   78.54 $\pm$ 1.09            &   81.78 $\pm$ 0.68                  \\ 
$l_2$-inv       &   78.81 $\pm$ 1.59        &   74.95 $\pm$ 2.32       \\ 
$l_0$-inv       & 73.68 $\pm$ 1.02          &   76.20 $\pm$ 1.95         \\ 
Badnets-SQ      &  70.85 $\pm$ 0.28               &   75.69 $\pm$ 0.14           \\ 
Badnets-PX      & 62.75  $\pm$  0.51                &   62.84 $\pm$ 0.26    \\ \hline
\end{tabular}
\begin{tablenotes}
\scriptsize
\item \textbf{Note:} The values represent mean $\pm$ standard deviation over three random seeds. \textit{Train} $\rightarrow$ \textit{Test}.
\end{tablenotes}
\end{threeparttable}
\end{table}

\subsection{Experimental Results}
To compare the performance of our methods in detecting unseen backdoored images, we train three widely used CNN architectures: Simple-CNN \cite{eykholt2018robust} consists of three convolutional layers and a fully connected, Deep-CNN \cite{stallkamp2012man} consists of six convolutional layers with a dropout layer and fully connected layer for classification, and ResNet-18 \cite{he2016deep}.

Table \ref{tab:main_results} presents the evaluation results, showcasing the exceptional performance of our proposed method in classifying unseen backdoor attack images. For example, on the CIFAR-10 dataset, our method achieves detection accuracies exceeding 95\% for Trojan-WM, Trojan-SQ, and \textit{l0}-inv triggers, alongside an impressive 93.85\% accuracy for \textit{l2}-inv triggers. Furthermore, the results highlight a 25\% increase for detecting Badnets-SQ triggers and over 4.5\% on Badnets-PX. To further validate the robustness of our approach, we extend our experiments to the GTSRB dataset. While prior CNN-based methods show promise in detecting Trojan triggers, our proposed method exhibits a remarkable performance increase, achieving an accuracy improvement of approximately 11\% and 8\% on Trojan-WM and Trojan-SQ, respectively. Additionally, our method significantly enhances detection accuracy by 35\% for unseen Badnets-SQ triggers on the GTSRB dataset. While our method performs well across most unseen attack types, the lower accuracy on Badnets-PX attacks highlights a limitation in detecting subtle, pixel-level triggers. The minimal changes may not significantly alter the global image features captured by the frozen visual encoder, making them more challenging to detect. Overall, these results validate the value of our approach in detecting unseen backdoor attacks across both datasets.

\subsection{Cross-Generalization Experiment}

Ensuring robust generalization across datasets is crucial for backdoored image detection, particularly when facing unseen triggers. To evaluate the strength of our proposed approach regardless of the dataset used during the training, we conduct experiments to train on one dataset (e.g. CIFAR-10) and test on the other (e.g. GTSRB), while ensuring that the model is still trained on seen triggers and tested on unseen triggers. Table \ref{tab:cross-gen} illustrates the impressive results of these experiments. For instance, in the initial tests (CIFAR-10 $\rightarrow$ GTSRB), our model achieves an average accuracy of 77.54\% on unseen Trojan triggers. The model retains robust performance across unseen \textit{l2}-inv and \textit{l0}-inv and triggers, achieving accuracies of 78.81\% and 73.68\%, respectively. 

When reversing the train and test sets (GTSRB $\rightarrow$ CIFAR-10), Trojan triggers achieve a higher average accuracy of 81.01\%. In this scenario, the model once again performs well in identifying unseen \textit{l2}-inv and \textit{l0}-inv and triggers. Interestingly, the model detects 62.84\% of unseen Badnets-PX triggers, which outperforms the performance when training directly on CIFAR-10. This improvement likely occurs due to the greater diversity and visual complexity of the GTSRB dataset (\textit{i.e.} 43 classes) compared to less diverse CIFAR-10 dataset (\textit{i.e.} 10 classes). This diversity appears to enable the model to learn more generalized representations, improving its ability to detect subtle pixel-level triggers like Badnets-PX.

\begin{figure}[t!]
\centering
\includegraphics[width=0.8\columnwidth]{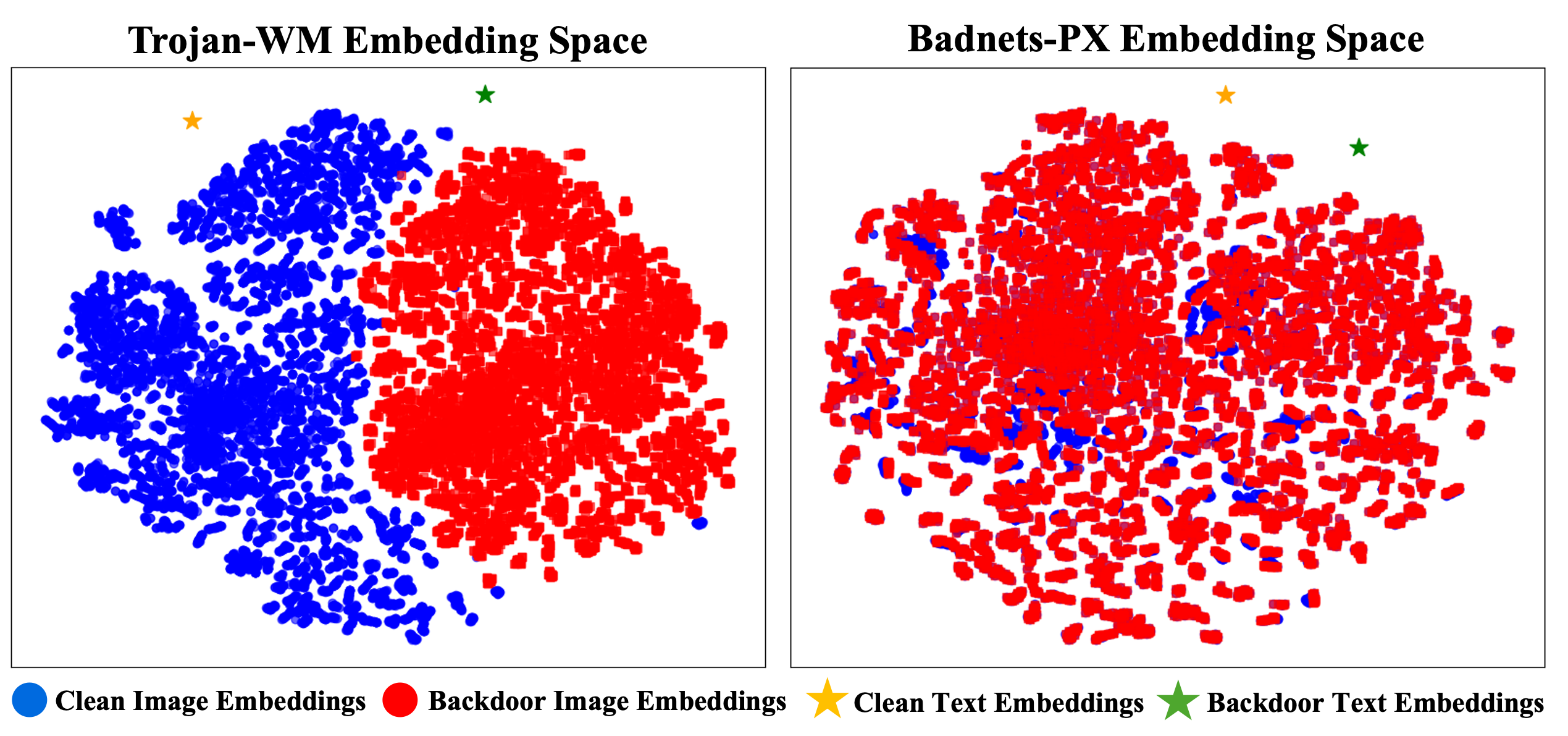}
\caption{t-SNE Visualization of test embeddings of Trojan-WM and Badnets-PX attacks on the CIFAR-10 dataset.}
\label{fig:tsne}
\end{figure}

\subsection{Visual Analysis}
To demonstrate the separation between clean and adversarial embeddings, we present t-SNE visualizations \cite{van2008visualizing} of the test image and text embeddings within the embedding space for unseen backdoor triggers Trojan-WM and Badnets-PX, shown in Figure \ref{fig:tsne}. These attacks were selected for illustration due to their contrasting detection performances: Trojan-WM achieves over 96\% accuracy, while Badnets-PX achieves around 59\%. For Trojan-WM, the correct text embeddings are closely aligned with their corresponding image embedding clusters, helping in create a distinct separation between clean and backdoor images. In contrast, the embedding space for Badnets-PX reveals a less distinct clustering pattern. While some separation occurs between the text embeddings, there is a significant overlap in image embeddings. This misalignment makes it more challenging to distinguish the unseen Backdoor-PX images from clean images during inference.

\subsection{Ablation on Fixed Prefix}

While our approach is built on leveraging a learnable prefix to adapt the textual representation in detecting unseen backdoored images, it is important to examine the effect on the model if the prefix remains static. We compare the results when using a fixed prompt -- ``a photo of'' -- with the performance using our learned prefix, shown in Table \ref{tab:static}. When the prompt remains static, the model relies heavily on the understanding of the fixed prefix, providing no additional semantic context that highlights backdoored features. Furthermore, the base model is applying only pre-trained knowledge without any fine-tuning, leading to a lack of generalization to previously unseen images of backdoored images. However, the learnable prefix helps guide the model's attention by enabling the prompt to dynamically adapt to associated backdoor patterns. This helps align visual and textual embeddings in the multimodal embedding space, making it more effective at detecting unseen backdoor triggers. 


\begin{table}[t!]
\centering
\scriptsize
\resizebox{\columnwidth}{!}{%
\begin{threeparttable}
\caption{Learnable vs. Static Prefix (Accuracy).}
\label{tab:static}
\begin{tabular}{|l|c|c|c|c|}
\hline
\multirow{2}{*}{\textbf{Unseen Attack}} & 
\multicolumn{2}{c|}{\textbf{CIFAR-10}} & 
\multicolumn{2}{c|}{\textbf{GTSRB}} \\ \cline{2-5}
 & \textbf{[p1][p2][p3]} & \textbf{``a photo of''} & \textbf{[p1][p2][p3]} & \textbf{``a photo of''} \\ \hline
Trojan-WM       &  96.10 (\textcolor{blue}{+42.87}) &  53.23  & 94.89 (\textcolor{blue}{+22.28}) & 72.61 \\ 
Trojan-SQ       &  96.44 (\textcolor{blue}{+43.14}) &  53.30  & 95.72 (\textcolor{blue}{+28.46}) & 67.26 \\ 
$l_2$-inv       &  93.85 (\textcolor{blue}{+39.84}) &  54.01  & 94.41 (\textcolor{blue}{+20.30}) & 74.11 \\ 
$l_0$-inv       &  96.45 (\textcolor{blue}{+44.06}) &  52.39  & 86.99 (\textcolor{blue}{+22.41}) & 64.58 \\ 
Badnets-SQ      &  75.45 (\textcolor{blue}{+24.14}) &  51.31  & 85.03 (\textcolor{blue}{+29.90}) & 55.13 \\ 
Badnets-PX      &  58.89 (\textcolor{blue}{+8.99})  &  49.90  & 60.42 (\textcolor{blue}{+8.83})  & 51.59 \\ \hline
\end{tabular}
\begin{tablenotes}
\scriptsize
\item \textbf{Note:} Differences in parentheses represent the accuracy improvement of the learned prefix over the static prefix. Values are shown in \textcolor{blue}{blue}.
\end{tablenotes}
\end{threeparttable}%
}
\end{table}

\section{Conclusion}

Defending object recognition systems against adversarial attacks has traditionally centered on reactive strategies like cleansing backdoored models or adversarially training them. In this groundbreaking work, we introduced a paradigm shift: a proactive method for detecting unseen backdoored (poisoned) images before they can infiltrate object recognition systems. Our approach serves dual purposes—vetting training datasets to ensure integrity and safeguarding inference by blocking adversarial images before they reach the model. We achieved this by harnessing the unparalleled generalization capabilities of vision-language models like CLIP, leveraging prompt tuning to exploit their training on vast and diverse datasets. Extensive experiments across six distinct types of unseen attacks demonstrate the robustness and effectiveness of our approach, setting a new benchmark for proactive defense mechanisms. While this pioneering work represents the first step toward detecting backdoored images, future research must delve deeper into improving detection of pixel-based attacks, where subtle, localized triggers present a formidable challenge. This study paves the way for a new era in securing object recognition systems against adversarial threats.  



%



\ifCLASSOPTIONcaptionsoff
  \newpage
\fi



%

\bibliographystyle{IEEEtran}
\bibliography{ref} 

%





\end{document}